\documentclass{article}
\usepackage{spconf}
\usepackage{amsmath,graphicx}
\usepackage{hyperref}    
\usepackage{url}            
\usepackage{booktabs}       
\usepackage{amsfonts}       

\usepackage{amsmath}
\usepackage{graphicx}
\usepackage[misc]{ifsym}

\newcommand{\vf}{{\boldsymbol{f}}}

\newcommand{\vx}{{\mathbf{x}}}

\newcommand{\vy}{{\mathbf{y}}}

\newcommand{\vs}{{\mathbf{s}}}
\newcommand{\bbR}{{\mathbb{R}}}

\newcommand{\gp}{{\mathcal{GP}}}
\newcommand{\N}{{\mathcal{N}}}
\newcommand{\F}{{\mathcal{F}}}

\newcommand{\freq}[1]{\hat{#1}} 
\newcommand{\timedelay}{{\theta}}
\newcommand{\phasedelay}{{\phi}}

\newcommand{\Var}{{\Sigma}}

\newcommand{\tra}{^{\top}}
\newcommand{\vmu}{{\boldsymbol{\mu}}}
\newcommand{\vtime}{{\boldsymbol{\timedelay}}}
\newcommand{\vphase}{{\boldsymbol{\phasedelay}}}

\newcommand{\gcsm}{\text{GCSM}}

\newcommand{\mtgp}{\text{M}}

\newcommand{\cov}{\text{cov}}
\newcommand{\cc}{\text{GCSM-CC}}
\newcommand{\sm}{\text{SM}}
\newcommand{\sg}{\text{SG}}
\newcommand{\itimej}[2]{{{#1}\times{#2}}}
\newcommand{\std}[1]{{\scalebox{0.9}{\hspace{1mm}$\pm$#1}}}

\name{Author(s) Name(s)  \thanks{Author Affiliation(s).}}
\address{Author Affiliation(s)}

\title{Multitask Gaussian Process with Hierarchical Latent Interactions}
\name{Kai Chen$^{\star \dagger}$\qquad Twan van Laarhoven$^{\dagger}$\qquad Elena Marchiori$^{\dagger}$\qquad Feng Yin$^{\star}$ (\Letter)\qquad Shuguang Cui$^{\star}$ }
\address{$^{\star}$Future Network of Intelligence Institute, The Chinese University of Hong Kong, Shenzhen, China.\\
$^{\dagger}$Institute for Computing and Information Sciences, Radboud University, Nijmegen, The Netherlands.
}

\begin{document}
    \maketitle
    \begin{abstract}        
        Multitask Gaussian process (MTGP) is powerful
        for joint learning of multiple tasks with complicated correlation patterns. However, due to the assembling of additive independent latent functions, all current MTGPs including the salient linear model of coregionalization (LMC) and convolution frameworks cannot effectively represent and learn the hierarchical latent interactions between its latent functions. In this paper, we further investigate the interactions in LMC of MTGP and then propose a novel kernel representation of the hierarchical interactions, which ameliorates both the expressiveness and the interpretability of MTGP. Specifically, we express the interaction as a product of function interaction and coefficient interaction. The function interaction is modeled by using cross convolution of latent functions. The coefficient interaction between the LMCs is described as a cross coregionalization term. We validate that considering the interactions can promote knowledge transferring in MTGP and compare our approach with some state-of-the-art MTGPs on both synthetic- and real-world datasets.      
    \end{abstract}

\begin{keywords}
    Gaussian process, multitask learning, linear model of coregionalization, latent interaction, spectral mixture kernel
\end{keywords}

\setcounter{section}{0}
\section{Introduction}   
Gaussian process (GP) \cite{Rasmussen2006} is an extraordinary Bayesian nonparametric model for representing the underlying function of a complex system due to its powerful fitting ability \cite{xu2019wireless}.  
The inferred GP model provides a posterior distribution over the underlying function, which can be refined as evidence is accumulated. 
The outstanding data fitting performance as well as its natural uncertainty quantification make GP competent in various sectors, for instance 6G wireless communication system \cite{9247529,9250516}.
However, the choice of kernel is a cornerstone for GP and influences the profile of its posterior distribution. 
The extension of GP to multiple correlated tasks is known as multitask (MT) GP (MTGP) \cite{bonilla2008multi}, which impressively shows the representation power in joint MT learning. MTGPs not only model nonlinear correlation among infinitely many random variables, as single task GPs (STGPs), but also account for high level correlation across different tasks  \cite{bonilla2008multi,Duerichen2015a,alvarez2012kernels}. For designing a MTGP, a very finicky problem is how to choose an expressive MTGP kernel, $k_{\mtgp}$, to jointly represent the cross covariance between tasks and auto-covariance within each single task \cite{Liu2018,alvarez2012kernels}. Given an expressive kernel, a MTGP can leverage knowledge from all tasks to obtain higher prediction accuracy than STGP independently leaning on each single task \cite{futoma2017learning,alaa2017deep}.

Early approaches to MTGPs, like Linear Model of Coregionalization (LMC)  \cite{bonilla2008multi,Goovaerts1997,Duerichen2015a}, focused on linear combination of independent latent functions represented by STGPs. More advanced MTGPs like the multi-kernel \cite{Melkumyan2011} and latent function convolution frameworks \cite{Alvarez2011,Guarnizo2018} adopt convolution to construct cross-covariance structure and assume that each task has its own auto-covariance structure. Comparing among the existing MTGPs, the LMC framework has eye-catching qualities, such as compact hyper-parameter space and interpretable hierarchical covariance structures. Since the introduction of the expressive and generalized spectral mixture (SM) kernels \cite{Wilson2013}, the learning capacities of MTGPs, including the LMC and convolution frameworks, have been ameliorated by incorporating SM kernels during the past years \cite{Ulrich2015a,Parra2017,chen2019multioutput}. However, these MTGPs have a generalized form with a sum of additive independent latent functions and dismiss the underlying interactions between the latent functions. 
In this paper, we aim at investigating and embodying interaction particularly for LMC due to its popularity.
Our work also provides an interesting cue for other MTGP frameworks, such as the convolution framework \cite{Melkumyan2011,Parra2017,chen2019multioutput}.

\textbf{Contributions:} We develop a novel LMC framework with hierarchical interactions for MTGP and the new features are as follows:
\begin{itemize}
        \item By using convolution between latent functions, we offer a kernel encoding function interactions in MTGP for the first time;
        
        \item We introduce cross coregionalization (CC) between Cholesky factors of LMCs to represent coefficient interactions of LMC; 
        
        \item We propose a kernel framework based on both the function and coefficient interactions for MTGP, which demonstrates rich representation, interpretability, and expressiveness.
            
        \item We investigate the learning capacity of the framework and collate its performance with recent competent MTGPs.
\end{itemize}
    
For the rest of the paper, GPs and related MTGPs are reviewed in Section \ref{bk}. In Section \ref{our}, our framework for MTGP is introduced and compared with some existing methods. Experiments on both synthetic- and real-world data are discussed in Section \ref{exp}. Finally, we conclude our work 
and touch upon some future investigations in Section \ref{sec:conclude}.

\section{Background and Related work}\label{bk}
\textbf{Gaussian process and kernel function:}
A GP defines a distribution over functions, specified by its mean function $m({\vx})$ and covariance function $k({\vx}, {{\vx}{'}})$ \cite{Rasmussen2006} for given input vector ${\vx}\in{\bbR}^{P}$. Mathematically, it is denoted  as $f\sim{\gp}(m({\vx}), k({\vx}, {{\vx}{'}}))$.
By placing a GP prior over functions through the choice of a kernel and hyper-parameter initialization, from the training data $\{ \mathbf{x}_i, y_i \}_{i=1}^{n}$ of finite size $n$, we can predict the unknown function value $\tilde{y}_*$ and its variance $\mathbb{V}[{y_*}]$ (that is, its uncertainty) for a test point ${\vx}_*$ using Bayesian inference.
The smoothness and generalization properties of a GP depend on the kernel function and its hyper-parameters ${\Theta}$. Consequently, various kernels, e.g. the squared exponential (SE) and kernel design methods have been introduced \cite{Rasmussen2006}. Based on the Bochner's theorem, the recent SM kernel have been proposed in \cite{Wilson2013}, which is derived through modeling spectral densities $\freq{k}_{\sm}(\vs)$ (Fourier transform of a kernel) with a Gaussian mixture. A desired property of SM kernel is that it can generalize any stationary kernel. The SM kernel has a form as $k_{\sm}(\tau)=\sum_{i=1}^{Q}k_{\sm, i}(\tau)$, 
where $k_{\sm, i}(\tau)={w_i}\exp(-2\pi^2\tau{\Var}_{i}\tau\tra)\cos(2\pi\vmu_{i}\tau\tra)$, $\tau={\vx}-{{\vx}{'}}$, $Q$ is the number of components, 
    $w_i$, $\vmu_{i}=[\mu_{i}^{(1)},...,\mu_{i}^{(P)}]$, and ${\Var}_{i}=\text{diag}([({\sigma_{i}^{2}})^{(1)},...,({\sigma_{i}^{2}})^{(P)}])$ are the weight, mean, and variance of the $i$-th Gaussian $\N_{i}(\vs; \vmu_{i}, {\Var}_{i})$ in frequency domain, respectively.
\begin{figure}[h!]
    \centering
    \renewcommand{\tabcolsep}{2.0mm}
    \begin{tabular}{p{2.0mm}*{2}{c}}
        & \includegraphics[width=0.29\columnwidth]{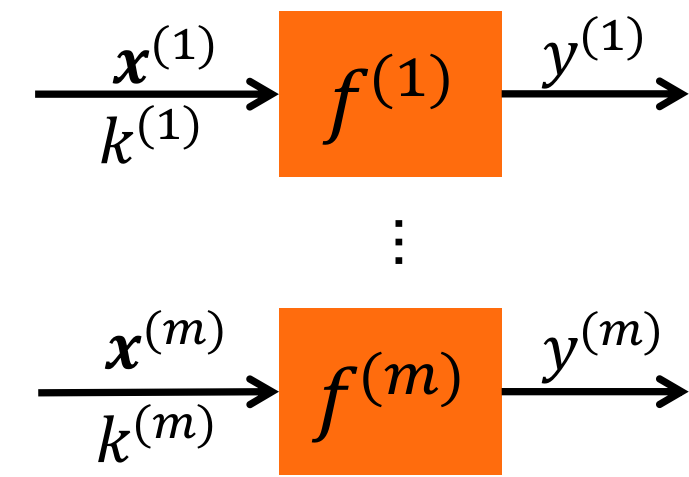} 
        & \includegraphics[width=0.27\columnwidth]{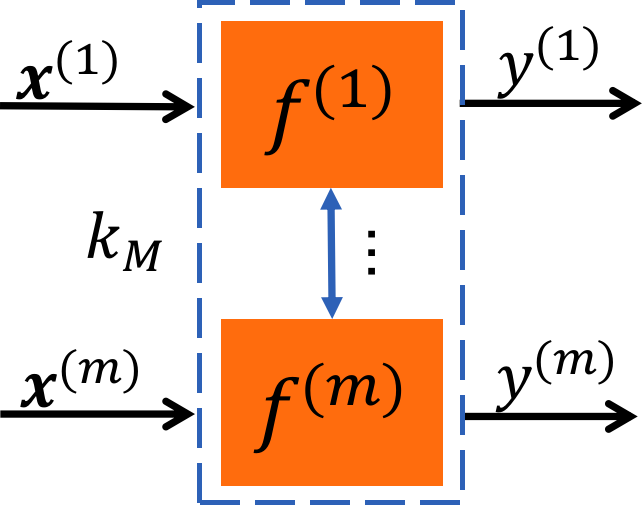}\\
        & {(a) $m$ STGPs with $m$ tasks} & {(b) A MTGP with $m$ tasks} 
    \end{tabular}
    \caption{The framework of STGP (a) and MTGP (b). }
    \label{fig:mtgp}
\end{figure}

\textbf{Multitask Gaussian process:} To clarify the mechanism of MTGP, we depict the conception of STGP and MTGP in Fig. \ref{fig:mtgp}. The GP model designated to the $m$-th task shown in subplot (a) has functional expressions: $f^{(m)}\sim{\gp}(0, k^{(m)}(\vx^{(m)}, \vx^{(m')}))$ and 
$y^{(m)}=f^{(m)}+\epsilon$, 
where $\epsilon$ is the noise of the model. The black arrow denotes data flow and inference direction. The orange rectangle denotes the underlying function of each task.     
For $m$ single tasks, each task modeled as an STGP and no correlation between tasks is taken into account. For MTGP shown in subplot (b), we wish to learn the $m$ underlying functions simultaneously with $\vf\sim\gp(0, k_{\mtgp}(X, X'))$ and $\vy=\vf+\boldsymbol{\epsilon}$, where $\vf=[f^{(1)}(\vx^{(1)}),...,f^{(m)}(\vx^{(m)})]^{\top}$, $X=[\vx^{(1)},...,\vx^{(m)}]^{\top}$, $\vy=[y^{(1)},...,y^{(m)}]^{\top}$, and $k_{\mtgp}(X, X')$ describes both the auto-covariance of each task and the cross-covariance between tasks. The dashed box in subplot (b) denotes the target MTGP model representing a joint Gaussian distribution of $m$ underlying functions. Hence, $k_{\mtgp}$ determines the learning capacity of MTGP. 

\textbf{Multitask kernel function:} The construction of a MTGP mainly involves
the designation of $k_{\mtgp}(X, X')$. Some state-of-the-art works on MT kernel design include: free form LMC first appeared in \cite{bonilla2008multi}, multi-kernel method \cite{Melkumyan2011}, GP regression network (GPRN) \cite{alaa2017deep,wilson2012gaussian}, Asymmetric focused MTGP \cite{Leen2012}, and multitask SM kernels \cite{Ulrich2015a,Parra2017,chen2019multioutput}. The general formula of $k_{\mtgp}$ can be written as matrix $K_{\mtgp}(X, X')={\begin{bmatrix}K^{(1,1)}, &K^{(1,m)} \\
        K^{(m,1)}, &K^{(m, m)}
        \end{bmatrix}},$
where the submatrix $K^{(m, 1)}$ describes the cross covariance between $f^{(m)}$ and $f^{(1)}$. An expressive $k_{\mtgp}$ using LMC linearly combines a mixture of $Q$ covariance components to ameliorate the representation of MTGP with multiple latent functions, which has an eventual form: ${K}_{\mtgp}=\sum_{i=1}^{Q}{B_{i}}\otimes{K_{s, i}}$, where $B_{i}$ is a LMC matrix representing task correlation and $K_{s, i}$ is a matrix constructed by arbitrary kernel of STGP. Since the introduction of SM kernels \cite{Wilson2013, Wilson2014a}, various MTGPs have been developed \cite{wilson2012gaussian,Ulrich2015a,Parra2017,chen2019multioutput} by substituting $k_{i}$ with $k_{\sm}$ or by building $k^{(m, 1)}$ as a convolution form.       
Due to the neat generalization and interpretability of SM, here we mainly review MTGPs incorporating SM.
Firstly, the cross SM (CSM) kernel \cite{Ulrich2015a} improved the expressiveness of GPRN: it contains cross phase spectrum and is also defined in a LMC form with ${K}_{\mtgp}=\sum_{i=1}^{Q}{B_{i}}\otimes{k_{\sg, i}}(\tau;{\Theta}^{i})$,
where ${k_{\sg, i}}(\tau;{\Theta}^{i})$ is phase notation of a spectral Gaussian kernel. The multi-output SM (MOSM) kernel \cite{Parra2017} provides a principled matrix factorization way to construct multivariate covariance functions with a better interpretation of the correlation between tasks. 
Even if MOSM extends existing MTGPs in expressiveness and interpretation, it still considers linear combination of latent functions and ignores interactions between them. By considering the imperfection of compatibility in MOSM when $m=1$, the more recent multi-output convolution SM (MOCSM) \cite{chen2019multioutput} employs both SM kernel and convolution mechanism and gain a more compatible MTGP.

\begin{figure}[h!]
    \centering
    \renewcommand{\tabcolsep}{1.0mm}
    \begin{tabular}{p{2.0mm}*{2}{c}}
        & \includegraphics[width=0.26\columnwidth]{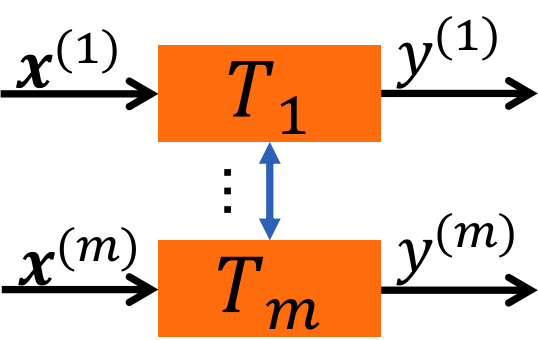} 
        & \includegraphics[width=0.32\columnwidth]{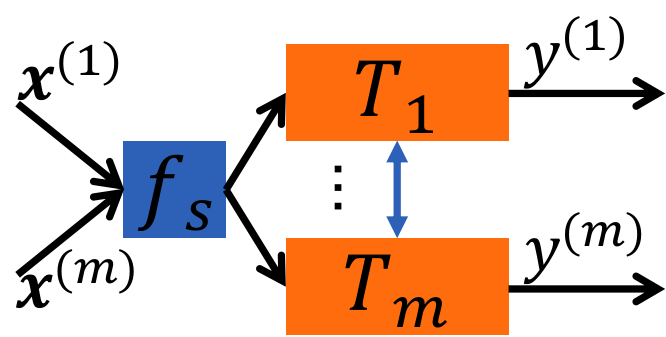}\\
        & {(a) MTGP using convolution} & {(b) MTGP using LMC}
    \end{tabular}
    \caption{Convolution (a) and LMC (b) frameworks of MTGP.}
    \label{fig:mtgp-lmc}
\end{figure}

\textbf{Comparison between MTGPs:}
In Fig. \ref{fig:mtgp-lmc}, we present and compare the two main frameworks of MTGPs: MTGPs using (a) convolution and (b) LMC. Here, $T_{m}$ labeled in the orange rectangle denotes the $m$-th task. In subplot (a), task correlation in MTGP is described by the convolution between latent functions of tasks and therefore each task has its own kernel function.  The bidirectional blue arrow denotes task correlation based on the convolution. In subplot (b), task correlation in LMC is hard-coded in a symmetrical hyper-parameter matrix and the $\{m, m'\}$-th entry of the matrix denotes the cross covariance between a pair of tasks $m, m'$. Diagonal elements in the matrix encode the auto-covariances and signal magnitudes of tasks. The differences between tasks are accounted by their different signal magnitudes. In particular, all tasks share a latent function space $f_{s}$ (denoted by blue rectange) adumbrating their common qualities. The shared $f_{s}$ means these tasks employ the same shared kernel for knowledge transferring and sharing. From Fig. \ref{fig:mtgp-lmc}, the LMC has a clearer hierarchical architecture for model explanation. In particular, the LMC has more compact hyper-parameter space than the convolution framework due to the shared latent function.
The architecture of LMC is simple and popularly used for many applications \cite{Duerichen2015a,futoma2017learning,wen2020modeling,wu2020multi}. That's one of the reason we mainly investigate the interaction of LMCs.

\section{Multitask spectral mixture kernel with interactions}\label{our}
For our attention on the LMC, a more general form of underlying function in MTGP can be amply  expressed as linear combination of independent latent functions,
    \begin{align}\label{eq:mtgp-func}
        f^{(m)}(\vx^{(m)})=\sum_{i=1}^{Q}\alpha^{(m)}_{i}f_{s, i}(\vx^{(m)}),
    \end{align}
    where $\alpha^{(m)}_{i}$ is a scalar coefficient and the independent latent function $f_{s, i}(\vx^{(m)})$ is a zero mean GP with $\cov[f_{s, i}(\vx^{(m)}), f_{s, i}(\vx^{(m')})]=k_{s, i}(\vx^{(m)}, \vx^{(m')})$. 
The linear combination form of such separate functions holds an assumption that $f_{s, i}(\vx^{(m)}){\perp}f_{s, j}(\vx^{(m)})$ and $\cov[f_{s, i}(\vx^{(m)}), f_{s, j}(\vx^{(m')})]=0$ when $i\neq{j}$. As mentioned before, kernel $k_{s, i}$ ensures the generalization of $f_{s, i}(\vx^{(m)})$. Due to the expressiveness and generalization of SM kernel, we substitute $k_{s, i}$ with $k_{\sm, i}$ and therefore obtain a more expressive
$f^{(m)}(\vx^{(m)})=\sum_{i=1}^{Q}\alpha^{(m)}_{i}f_{\sm, i}(\vx^{(m)})$, 
where $f_{\sm, i}\sim\gp(0, k_{\sm, i})$. On the other hand,  for STGP using SM kernel, the dependency between latent functions $\{f_{\sm, i}, f_{\sm, j}\}$ has been investigated and proved by \cite{kai2019}, which is somewhat significant. However, the interaction between latent functions of aforementioned MTGPs is unclear and unexplained. Hence, in this paper, we consider two hierarchical interactions in LMC, function interaction between latent functions $f_{\sm, i}$ and $f_{\sm, j}$ and coefficient interaction between $\alpha^{(m)}_{i}$ and $\alpha^{(m)}_{j}$.

\begin{figure}[h!]
    \centering
    \renewcommand{\tabcolsep}{0.1mm}
    \def\figart#1{\includegraphics[width=0.33\columnwidth]{fi-#1-GCSM.pdf}}
    \begin{tabular}{p{1.0mm}*{3}{c}}
        & \figart{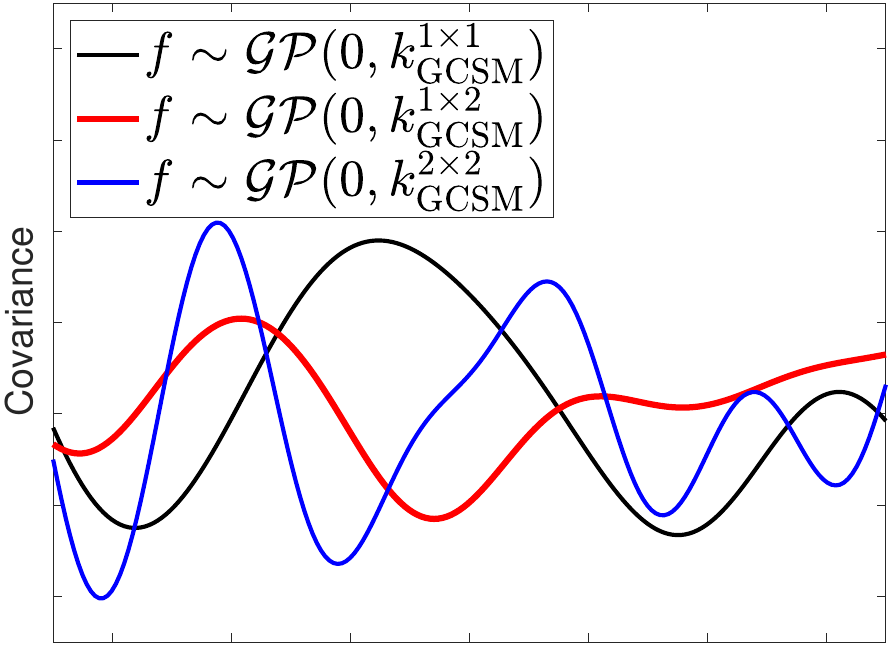}
        & \figart{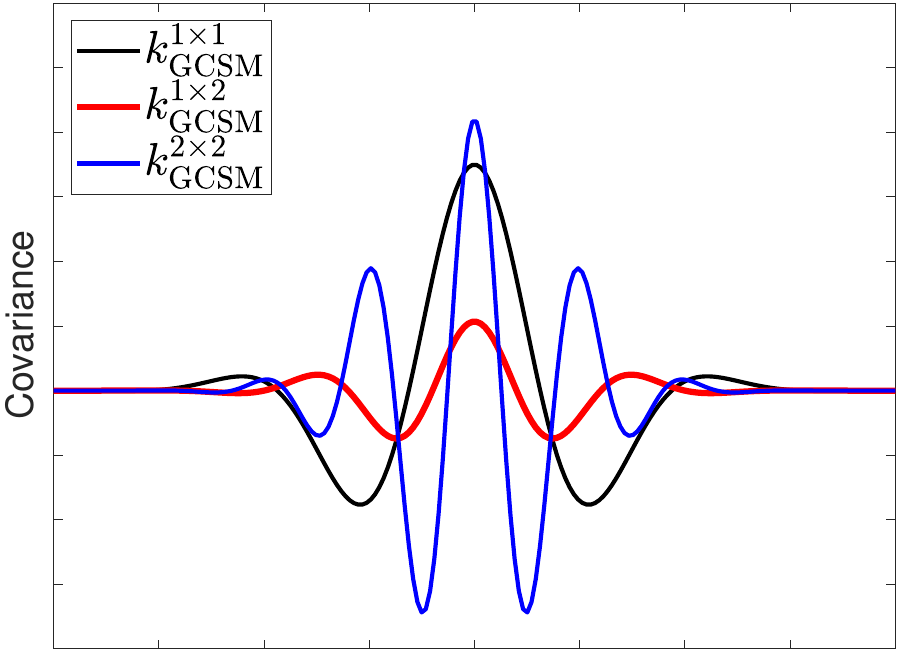}
        & \figart{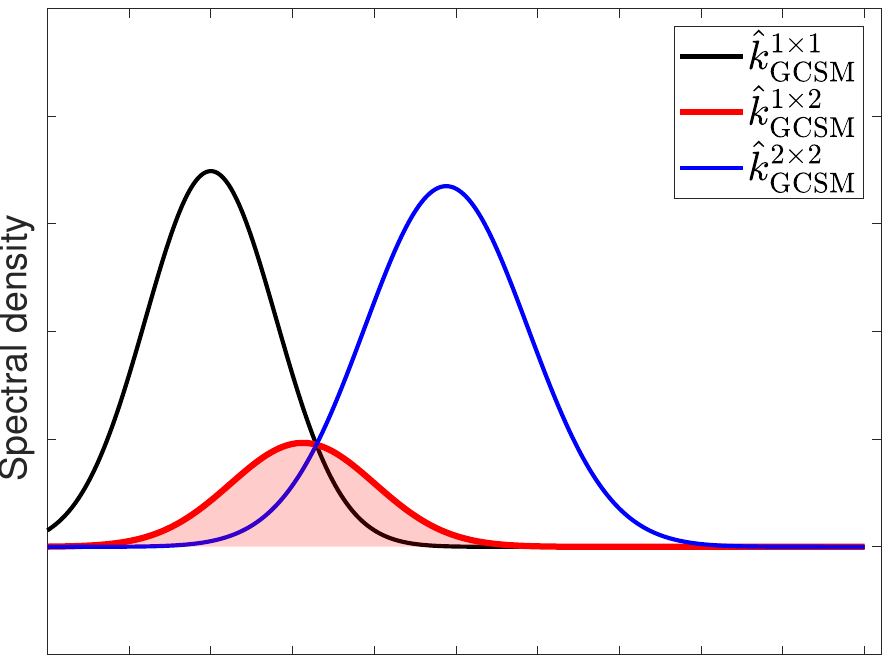}\\
        & (a) Latent functions & (b) The covariances & (c) The SDs 
    \end{tabular}
    \caption{The illustrations of latent functions (a), covariances (b), and SDs (c) of their interaction.}
    \label{fig:sm-dep}
\end{figure}

\textbf{Function interaction between latent functions:} 
In this paper, we follow the investigation in \cite{kai2019} and view the dependency as a general sense of function interaction between latent functions.
Furthermore, for $k_{\sm, i}$, we borrow the parameterization of time and phase delayed SM kernel in \cite{kai2021SMD,chen2019multioutput} to enrich the features of function interaction. Then, we meliorate the function interaction in LMC by describing it related to time and phase delay, as modeled in \cite{kai2019,kai2021SMD} through the so-called generalized convolution spectral mixture (GCSM) kernel, defined as follows:
    \begin{align}\label{eq:GCSM}
    \begin{split}
    k_\gcsm^{\itimej{i}{j}}(\tau)
    =&{\F}_{s\rightarrow \tau}^{-1}\big[{\freq{k}_\gcsm^{\itimej{i}{j}}({\vs})}+{\freq{k}_\gcsm^{\itimej{i}{j}}({-\vs})}\big](\tau)\\
    =&{c_{ij}}\exp\big(-\frac{1}{2}{\tau_{\vtime}\tra{{\Var}_{ij}}\tau_{\vtime}}\big)\cos\big( {\tau_{\vtime}\tra{\vmu}_{ij}}-{{\vphase}_{ij}\pi}\big),    
    \end{split}
    \end{align}
    where $\vs$ is a spectral vector, $c_{ij}=w_{ij}a_{ij}$ is the cross constant incorporating cross weight and amplitude, and $\tau_{\vtime}=2\pi(\tau-\frac{{\vtime}_{ij}}{2})$ is the Euclidean distance with time delay. Other parameters related to $\{f_{\sm, i}, f_{\sm, j}\}$ are defined as: cross weight, $w_{ij}=\sqrt{w_{i}w_{j}}$; cross amplitude, ${a}_{ij}={\big|4\pi^{2}\Sigma_{i}\Sigma_{j}\big|}^{\frac{1}{4}}\N(\vmu_{i};\,\vmu_{j}, \frac{\Sigma_{i}+\Sigma_{j}}{2})$; cross mean, $\vmu_{ij}=\frac {{\Var}_{i}{\vmu}_{j}+{\Var}_{j}{\vmu}_{i}}{{{\Var}_{i}+{\Var}_{j}}}$; cross length scale, ${{\Var}}_{ij}=\frac{{2{{\Var}_{i}{\Var}_{j}}}}{{{\Var}_{i}+{\Var}_{j}}}$; cross time delay, ${\vtime}_{ij}={\vtime}_{i}-{\vtime}_{j}$, and cross phase delay, ${\vphase}_{ij}={\vphase}_{i}-{\vphase}_{j}$,      
    where $\vtime_{i}$ and $\boldsymbol{\phi}_{i}$ respectively denote latent time and phase delays in $f_{\sm, i}$. Here, ${\freq{k}_{\gcsm}^{\itimej{i}{j}}({\vs})}$ is a cross spectral density (SD) between the $k_{\sm, i}$ and $k_{\sm, j}$,
    \begin{align}
    \begin{split}
    {\freq{k}_\gcsm^{\itimej{i}{j}}({\vs})}
    ={c_{ij}}\N\big(\vs;\vmu_{ij},\Sigma_{ij}\big)
    \exp\big(-\pi{\imath}({\vtime}_{ij}{\vs}+{{\vphase}_{ij}})\big),
    \end{split}
    \end{align}
    where the hat symbol denotes SD. 
    Note that ${k_\gcsm^{\itimej{i}{j}}}(\tau)=k_{\sm, i}$ and $\freq{k}_\gcsm^{\itimej{i}{j}}(\tau)=\freq{k}_{\sm, i}$ for $i=j$, where $\freq{k}_{\sm, i}$ is the SD of $k_{\sm, i}$. 
 The SD and covariance of the interaction (in red) between $f_{\sm,i}$ and $f_{\sm, j}$ are illustrated in Fig. \ref{fig:sm-dep}.
    Through the GCSM we can model the function interaction 
    between $\{f_{\sm, i}, f_{\sm, j}\}$.
    
\textbf{Coefficient interaction between LMCs: } For the coefficient interaction between $\alpha^{(m)}_{i}$ and $\alpha^{(m)}_{j}$, we reformulate $\alpha^{(m)}_{i}$ in a LMC matrix ${B}_{i}$ as $B_{i}(m, m')=\alpha^{(m)}_{i}\alpha^{(m')}_{i}$, where $B_{i}(m, m')$ is the element of $B_{i}$ located at the $m$-th column and $m'$-th row. In addition, we consider the free-form parameterization of ${B}_{i}$ in \cite{bonilla2008multi,Liu2018,alvarez2012kernels} and decompose ${B}_{i}$ using Cholesky factorization as: ${B}_{i}=B_{\textit{L}, i}B_{\textit{L}, i}^{\top}$ with the Cholesky factor, low triangle  $B_{\textit{L}, i}={\begin{bmatrix}\ell_{1,1}^{i}, & 0
        \\ \ell_{m,1}^{i}, &\ell_{m,m}^{i}
        \end{bmatrix}}$, 
where $\ell_{m, m'}^{i}$ can be seen as the correlation between tasks $m$ and $m'$ \cite{bonilla2008multi}. Note that there are in total ${m(m+1)}/{2}$ hyper-parameters for the free-form parameterization of ${B}_{i}$ in LMC. Furthermore, for any two arbitrary LMC terms ${B}_{i}$ and ${B}_{j}$, we construct ${B}_{ij}=B_{\textit{L}, i}B_{\textit{L}, j}^{\top}$ to encode the coefficient interaction between LMCs. Here we interpretate ${B}_{i,j}$  as a cross coregionalization (CC) term, which is capable of capturing complicated interaction between coefficients. There is no extra hyper-parameter introduced for the coefficient interaction. In particular, a compatibility is guaranteed such that ${B}_{ij}$ becomes $B_{i}$ when $i=j$. Thus, for MTGP with $Q$ latent functions constructed by using LMC and SM kernel, we have a kernel with hierarchical interactions as follows:
\begin{align}\label{eq:cc}
    \begin{split}
    K_{\cc}(\tau)
    &=\sum_{i=1}^{Q}\sum_{j=1}^{Q}{B}_{ij}\otimes{k_\gcsm^{\itimej{i}{j}}}(\tau).
    \end{split}
\end{align}
Due to the use of GCSM and CC, we call the proposed kernel as GCSM-CC. The positive semi-definite (PSD) of GCSM-CC is guaranteed by the LMC framework and the PSD of both $B_{i}$ \cite{bonilla2008multi} and $k_{\gcsm}$ \cite{kai2019}. Note that GCSM-CC is the first work of non-separable LMC. 
    
\textbf{The interpretation of interaction:}
In addition to existing MTGPs, the GCSM-CC has following desired properties: (1) it does not treat latent functions separately; (2) it allows to model function interaction including time and phase delay; (3) it uses CC to represent coefficient interaction across all LMCs; (4) it includes more interaction terms without the increasing of hyper-parameter space (if remove time and phase delays).      
As shown in Fig. \ref{fig:sm-dep}, there are two latent functions, $f\sim\gp(0, k_\gcsm^{\itimej{1}{1}})$ (in black) and $f\sim\gp(0, k_\gcsm^{\itimej{2}{2}})$ (in blue). From subplot (c), the SD $\freq{k}_\gcsm^{\itimej{1}{2}}$ of the interaction in the frequency domain can be seen as an intersection between $\freq{k}_\gcsm^{\itimej{1}{1}}$ and $\freq{k}_\gcsm^{\itimej{2}{2}}$, where $\freq{k}_\gcsm^{\itimej{1}{1}}=\freq{k}_{\sm, 1}$. The covariance ${k}_\gcsm^{\itimej{1}{2}}$ (subplot (b)) of the interaction in the time domain has a period smaller than $k_{\sm, 1}$ and bigger than $k_{\sm, 2}$. For the latent function $f\sim\gp(0, k_\gcsm^{\itimej{1}{2}})$, covariance $k_\gcsm^{\itimej{1}{2}}$, and SD $\freq{k}_\gcsm^{\itimej{1}{2}}$ of the interaction, their magnitudes are smaller than the original latent functions. 
Due to the distributivity and commutativity of interaction, there are $Q^2$ interaction terms with $Q^2-Q$ cross interaction terms plus $Q$ auto interaction terms in GCSM-CC. The cross interactions allow extensive communication between latent functions and hence bring potent fitting capacity.

\section{Experiments}\label{exp}
In this section, we compare GCSM-CC with some existing MTGPs  \cite{bonilla2008multi,wilson2012gaussian,Ulrich2015a,Parra2017}. First we show the ability of GCSM-CC to simultaneously model multiple incomplete signals sampled from a GP, its integral and derivative.
Then we use GCSM-CC for prediction tasks on a real-world problem with sensor signals \footnote{http://slb.nu/slbanalys/historiska-data-luft/} related to air pollution monitoring:  Nitrogen oxide (NO) concentration.  
For both the synthetic and real-world datasets, we follow the experimental setting in \cite{Alvarez2011,Parra2017} to generate incomplete signal. However, we observed that in the considered experimental setting it is not difficult to capture correlation between tasks because the training data in one signal has a consecutive intersection with the training data of another signal. In particular, we consider a more challenging task that does not involve consecutive intersection of training data between tasks and consider also a signal recovery. 
We implemented our models using GPflow \cite{Matthews2017} to improve scalability and facilitate gradient computation. The open source codes will be made available soon. 
We use the mean absolute error ${\mathrm {MAE} ={{\sum _{i=1}^{n}\big|y_{i}-\tilde{y}_{i}\big|}/{n}}}$ as performance metric.

\subsection{Synthetic experiment for symmetric predictions of all tasks}
    We conduct an synthetic experiment learning multiple incomplete signals. In this context,  we validate the interpolation, extrapolation, and signal recovery ability of GCSM-CC and compare its pattern recognition performance with that of other MTGPs. We consider three tasks  corresponding to three signals: a mixed signal, its integral, and its derivative, respectively.  Specifically, we generate a Gaussian signal with length 300 in the interval [-10, 10] and numerically compute its first integral and derivative. In this experiment, MTGPs allow bi-directional knowledge transfer between tasks to improve the symmetric predictions of all tasks. 
        
\begin{figure}[h!]
    \centering
    \renewcommand{\tabcolsep}{-1.0mm}
    \def\figart#1{\includegraphics[trim=10 5 5 5, clip, width=0.52\columnwidth]{fi_art_DSM_#1.png}}   
    \begin{tabular}{p{-2.0mm}*{3}{c}}
        & \figart{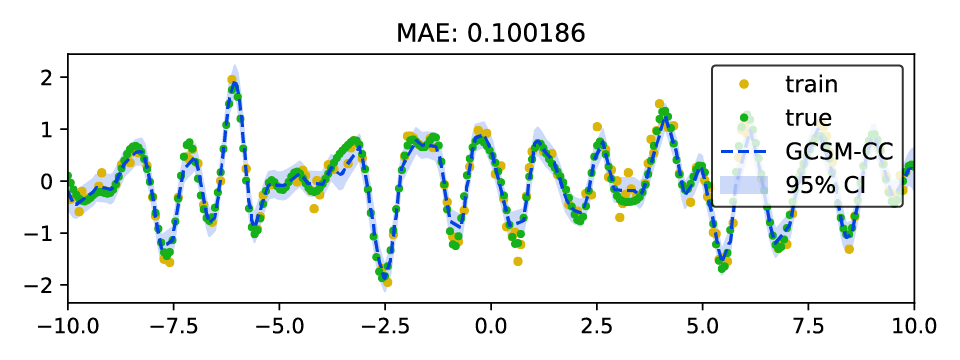} & 
        & \figart{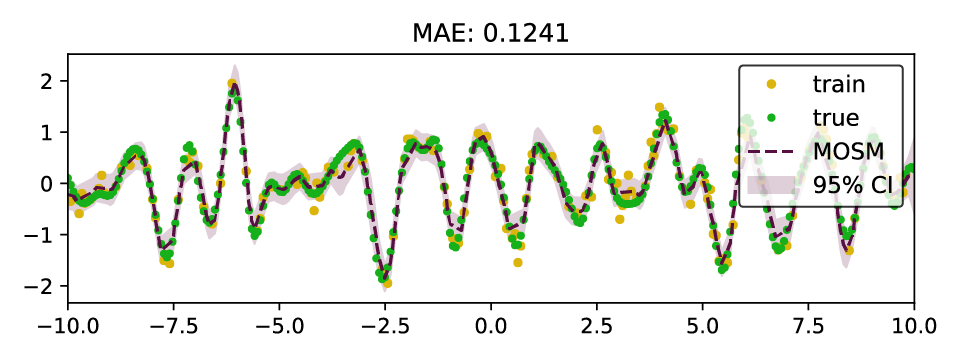}\\
        & & {(a)} & \\ 
        & \figart{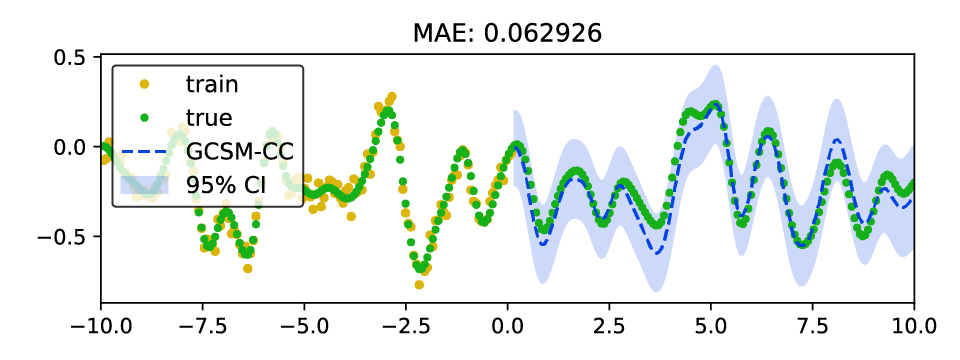} &
        & \figart{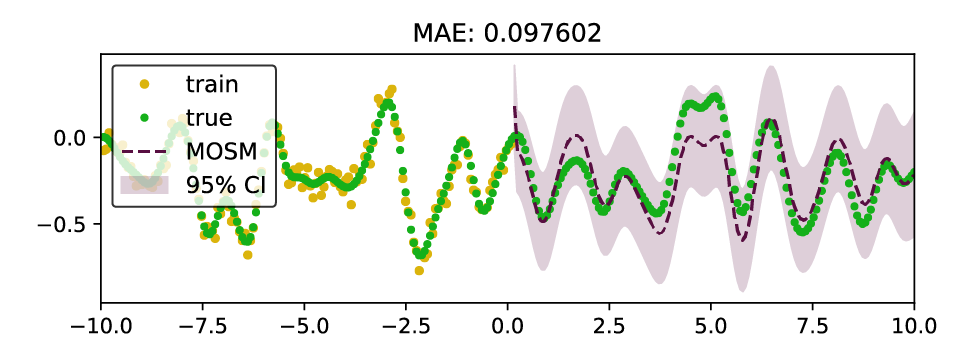}\\
        & & {(b)} &  \\
        & \figart{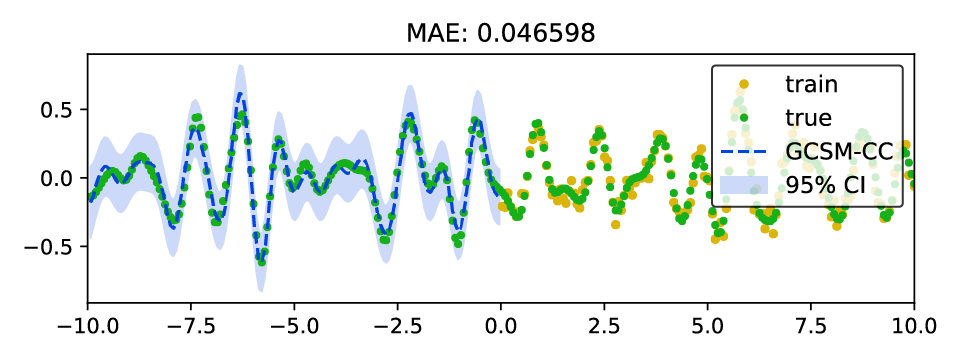} &
        & \figart{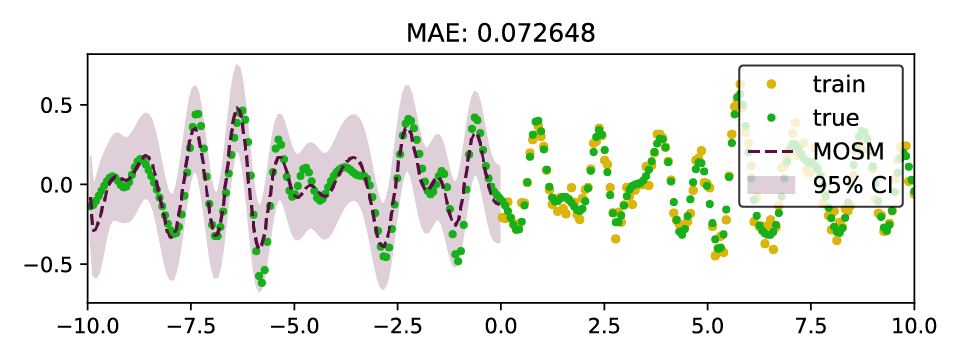} \\
        & & {(c)} & \\
    \end{tabular}
        \caption{Performance of GCSM-CC (in blue dashed line) and MOSM (in plum dashed line) on synthetic MT. Subplot (a): the signal of task 1 sampled from $f\sim\mathcal{GP}(0, K_\textit{SM})$. Subplot (b): task 2 with integral of the signal. Subplot (c): task 3 with derivative of the signal.}
\label{fig:arti}
\end{figure}

For the 1st task with a signal sampled from $\mathcal{GP}(0, K_\textit{SM})$,  we randomly choose half as the training data, and the rest as testing data. For the 2nd task, the integral of the signal in the interval [-10, 0] are used for training (in dark yellow), while the remaining signal points in the interval [0, 10] are used for testing (in green). For the 3rd task, the derivative of the signal in the interval [0, 10] is used for training and the rest for testing. The performance of GCSM-CC on the generated signal is shown in Figure \ref{fig:arti} (a). According to Table \ref{tab:t_exp1}, all considered MTGPs have comparable performance: they interpolate well the missing values. 
The second task, i.e., the integral of the signal, is shown in Figure \ref{fig:arti} (b). In this case its inherent patterns are more difficult to recognize and extrapolate. Here, GCSM-CC is superior to all other methods: it achieves the lowest MAE as well as the smallest confidence interval (CI). Besides, GCSM-CC excels on the last task, i.e., the derivative signal (see Figure \ref{fig:arti} (c)): it shows the best pattern learning and extrapolation capability while using  the same number of base component ($Q=10$). 
Overall, these results indicate that the outstanding capability of GCSM-CC in capturing integration and differentiation patterns of the generated signal simultaneously.

\begin{table}
    \tiny
    \caption{Performance (MAE) of GCSM-CC and other methods.}\label{tab:t_exp1}
    \setlength{\tabcolsep}{3.0pt}
    \begin{center}		
        \begin{tabular}{c r @{} l r @{} l r @{} l r @{} l r @{} l r @{} l r @{} l}
            \toprule
            {Signal} & \multicolumn{2}{c}{SE-LMC} & \multicolumn{2}{c}{Mat\'ern-LMC} & \multicolumn{2}{c}{GPRN} & \multicolumn{2}{c}{CSM} & \multicolumn{2}{c}{MOSM} & \multicolumn{2}{c}{GCSM-CC} \\
            \midrule
            
            Mixed signal & 0&.16$\std{0.01}$ & 0&.11$\std{0.01}$ & 0&.12$\std{0.01}$ & 0&.12$\std{0.01}$ & 0&.13$\std{0.01}$ & $\boldsymbol{0}$ & $\boldsymbol{.10}$$\std{0.003}$  \\
            
            Integral  & 0&.26$\std{0.01}$ & 0&.25$\std{0.02}$ & 0&.33$\std{0.05}$ & 0&.19$\std{0.06}$ & 0&.09$\std{0.004}$ & $\boldsymbol{0}$ & $\boldsymbol{.06}$$\std{0.003}$ \\
            
            Derivative  & 0&.18$\std{0.01}$ & 0&.19$\std{0.01}$ & 0&.09$\std{0.01}$ & 0&.17$\std{0.02}$ & 0&.08$\std{0.01}$ & $\boldsymbol{0}$ & $\boldsymbol{.04}$$\std{0.01}$ \\
            
            NO$^{H}$ & 130&.96$\std{0.41}$ & 132&.89$\std{0.37}$ & 58&.16$\std{1.17}$ & 52&.02$\std{4.28}$ & 53&.95$\std{1.04}$ & $\boldsymbol{41}$ & $\boldsymbol{.16}$$\std{0.95}$ \\
            
            NO$^{S}$ & 85&.06$\std{0.38}$ &  85&.19$\std{0.36}$ & 45&.98$\std{2.61}$ & 35&.48$\std{1.17}$ & 60&.81$\std{1.60}$ & $\boldsymbol{33}$ & $\boldsymbol{.39}$$\std{1.54}$ \\
            \bottomrule
        \end{tabular}
    \end{center}    
\end{table}

\subsection{Nitrogen oxides concentration for asymmetric extrapolations of primary tasks}

On the contrary, for this real-world experiment, MTGPs focus on improving the prediction performance of primary tasks by transferring valuable knowledge from other correlated tasks.
The sensor network dataset recorded from Stockholm city monitor air pollution parameters in order to provide air quality surveillance for the regional environment.  In particular, NO is an important parameter, since long term exposure at high concentration can cause inflammation of the human airways. Extrapolation and forecasting models allow to monitor NO concentration in order to control and prevent negative effects on health and environment. As the first real world dataset, we use NO concentration from 5 January, 2017 to 25 January, 2017, in one-hour interval, collected at three stations (Essingeleden, Hornsgatan, Sveav\"agen) in Stockholm. 
    
Each station corresponds to a single task: Essingeleden as task 1,  Hornsgatan as task 2 and  Sveav\"agen  as task 3. NO evolution shows time and phase related patterns and their variability over the period of recording. Different stations have different local patterns impacted by the station's surroundings. For instance: Essingeleden's measurement are recorded at open path, Hornsgatan's measurement and Sveav\"agen's measurement at street. Still, these tasks have shared global trends because of the global seasonal change and periodic characteristics of human and industry activities. The evolution of NO concentration in each task is a result of nonlinear interaction of time- and phase dependent local and global patterns. Therefore, knowledge from correlated tasks should help each other when used to model long range trends.

We aim to assess comparatively the long range extrapolation ability of GCSM-CC for future forecasting and signal recovery simultaneously. Therefore,  we perform extrapolation for primary tasks 2 and 3.
For task 1, we randomly chose half of the Essingeleden time series as the  training data. For task 2, the first half of the Hornsgatan time series is used for training and the remaining data for testing. For task 3, the last half of the Sveav\"agen time series  is used for training and the rest for testing. 
Table \ref{tab:t_exp1} reports the performance of the considered MTGPs on each task of the two experiments. Significantly, the GCSM-CC achieves the lowest MAE.
    
       
\begin{figure}[h!]
    \centering
    \renewcommand{\tabcolsep}{-0.2mm}
    \def\figreal#1{\includegraphics[trim=10 5 5 5, clip, width=0.52\columnwidth]{fi_art_DSM_SLB-NO_1-#1.png}}
    \begin{tabular}{p{-2.0mm}*{2}{c}}
        & \figreal{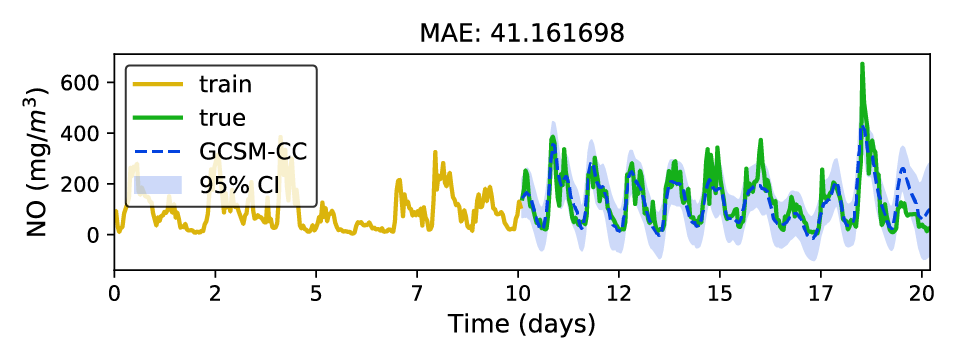} 
        & \figreal{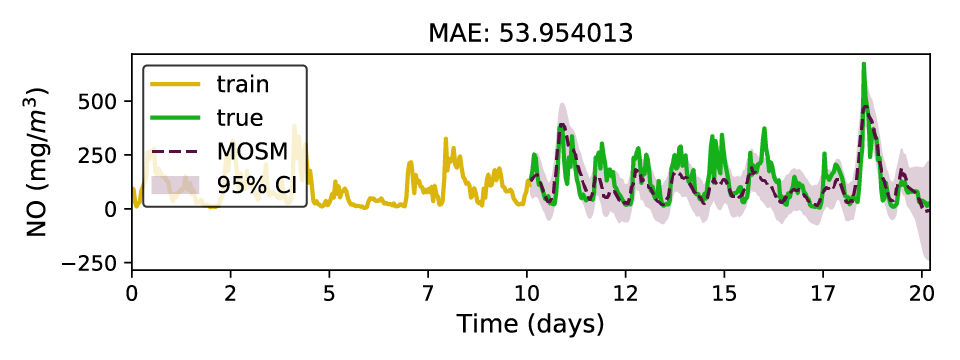}\\
        & {(a) GCSM-CC} & {(b) MOSM} \\ 
        & \figreal{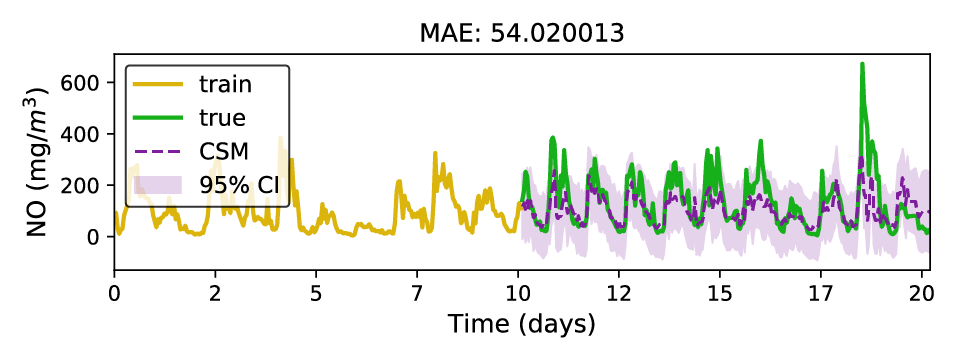} 
        & \figreal{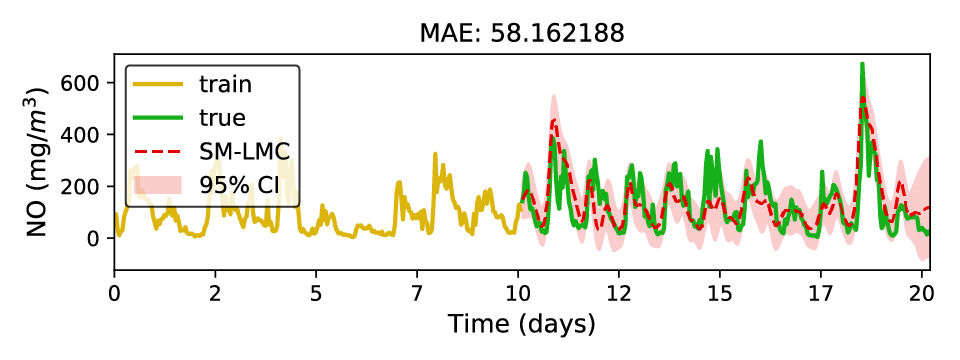}\\
        & {(c) CSM} & {(d) GPRN} 
    \end{tabular}
    \caption{NO concentration extrapolations. }\label{fig:real1}
\end{figure}
              
\section{Conclusion}\label{sec:conclude}
We have proposed the GCSM-CC kernel for MTGP with hierarchical interactions.
Hierarchical interactions include function interaction modeled by using the dependency between latent functions specified by SM kernel, coefficient interaction constructed by using cross coregionalization. In this way, GCSM-CC advances the learning capacity and interpretability of MTGPs beyond non-interactive frameworks. 
Interesting future research involves the development of sparse and efficient inference methods for this interactive MTGP  \cite{xie2019distributed,lin2019multiobjective}.
    
    
    

%


\bibliographystyle{IEEEbib}
\bibliography{main.bbl}
\end{document}